\documentclass[lettersize,journal]{IEEEtran}
\usepackage{amsfonts}
\usepackage{algorithmic}
\usepackage{algorithm}
\usepackage{array}
\usepackage{textcomp}
\usepackage{stfloats}
\usepackage{url}
\usepackage{verbatim}
\usepackage{graphicx}
\usepackage{subfigure}
\usepackage{cite}
\usepackage{booktabs}
\usepackage{color}
\usepackage{amssymb}
\usepackage[tbtags]{amsmath}
\usepackage[thmmarks,amsmath]{ntheorem}
\qedsymbol{\ensuremath{\square}}
{
	\theoremstyle{nonumberplain}
	\theoremheaderfont{\bfseries}
	\theorembodyfont{\normalfont}
	\theoremsymbol{\ensuremath{\square}}
	
}

\newtheorem{theorem}{\textbf{Theorem}}

\begin{document}
\title{Towards Carbon-Neutral Edge Computing: Greening Edge AI by Harnessing Spot and Future Carbon Markets}

\author{Huirong~Ma,
	Zhi~Zhou,~\IEEEmembership{Member,~IEEE,}
        Xiaoxi~Zhang,~\IEEEmembership{Member,~IEEE,}
	and~Xu~Chen,~\IEEEmembership{Senior Member,~IEEE}
 \thanks{H. Ma, Z. Zhou, X. Zhang and X. Chen are with the School of Computer Science and Engineering, Sun Yat-sen University, China. E-mail: \{mahr6, zhouzhi9, zhangxx89, chenxu35\}@mail.sysu.edu.cn.}
  }

\maketitle

\begin{abstract}
Provisioning dynamic machine learning (ML) inference as a service for artificial intelligence (AI) applications of edge devices faces many challenges, including the trade-off among accuracy loss, carbon emission, and unknown future costs.
Besides, many governments are launching carbon emission rights (CER) for operators to reduce carbon emissions further to reverse climate change.
Facing these challenges, to achieve carbon-aware ML task offloading under limited carbon emission rights thus to achieve green edge AI,
we establish a joint ML task offloading and CER purchasing problem, intending to minimize the accuracy loss under the long-term time-averaged cost budget of purchasing the required CER.
However, considering the uncertainty of the resource prices, the CER purchasing prices, the carbon intensity of sites, and ML tasks' arrivals, it is hard to decide the optimal policy online over a long-running period time.
To overcome this difficulty, we leverage the two-timescale Lyapunov optimization technique, of which the $T$-slot drift-plus-penalty methodology inspires us to propose an online algorithm that purchases CER in multiple timescales (on-preserved in carbon future market and on-demanded in the carbon spot market) and makes decisions about where to offload ML tasks. Considering the NP-hardness of the $T$-slot problems, we further propose the resource-restricted randomized dependent rounding algorithm to help to gain the near-optimal solution with no help of any future information.
Our theoretical analysis and {{extensive}} simulation results driven by the real carbon intensity trace show the superior performance of the proposed algorithms.
\end{abstract}


\section{Introduction}
\IEEEPARstart{W}{ith} the rapid development of machine learning (ML) technique and the spread of mobile and Internet of Things (IoT) devices running artificial intelligence (AI) applications, using Deep Neural Network (DNN) model to infer ML tasks becomes increasingly popular \cite{a03,a04,a0404}. However, devices usually have limited energy in the battery and are restricted by the physical size \cite{a040404}, thus failing to support the inference execution of ML tasks that need more computation resources. To efficiently compute ML tasks, one way is to utilize the resource-rich cloud server. However, transferring ML tasks through the wide area network (WAN) will incur extravagant energy consumption, let alone the expensive price of energy and high carbon intensity in the cloud server. Another way is offloading ML tasks to edge servers, which can take advantage of the green energy to achieve low-carbon computation at the network edge \cite{a01,a0101,a010101}. Unfortunately, the achieved inference accuracy using edge resources is usually worse than that in the cloud server owing to the limited resource. Hence, neither an edge-only nor a cloud-only solution is the best to support the low-carbon and high-accuracy ML inference services.
\begin{figure}[!t]
	\centering
	\includegraphics[width=0.999\linewidth]{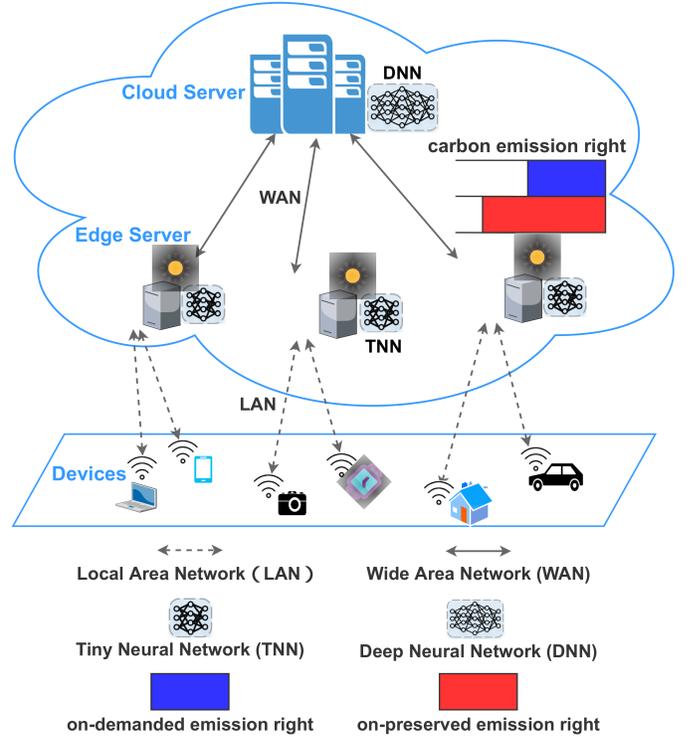}
	\caption{An illustration of the carbon-aware ML task offloading in the collaborative edge computing framework.}\vspace{-0.25cm}
\end{figure}

Additionally, to mitigate the problem of global warming incurred by carbon emissions, carbon emission rights (CER) trading (e.g., UN Carbon Offset Platform \cite{a05}), which can be viewed as an offsetting mechanism for trading carbon credits, has been adopted globally \cite{a0505}. As for a CER trading scheme, the purchaser needs to purchase CER from a central authority or government body, and can only discharge a specific quantity of pollutants during a specific time.
For example, there is a purchaser who is an operator owning multiple edge servers and a cloud server, and its' annual CER is 6 million tons. Therefore, the {{operator's}} all servers can only emit a total of 6 million tons of carbon per year. When the CER is not sufficient, the operator needs to purchase additional CER.
Usually, the operator can purchase CER in the following two ways, including but not limited to: one is purchasing CER in the spot market in a timely manner for what the system needs (on-demanded) in a shorter period (e.g., seconds or minutes); the other is to purchase in the carbon future market in advance (on-preserved) in a specified period (e.g., minutes or hours).
Obviously, for the same value of CER, the cost of the latter method is cheaper, which is consistent with the real situations in the market \cite{a07}. Usually, the operator has a limited budget to purchase CER, e.g. annual budget.
Therefore, a proper policy that can be integrated with the above two methods is needed to obey the CER purchasing budget.

Carbon-aware ML task offloading in the collaborative edge computing framework, illustrated in Fig. 1, which coordinates resource-constrained but carbon-intensity lower edge servers and the resource-rich but carbon-intensity higher cloud server, will provide low-carbon and high-accuracy inference services.
In this framework, ML tasks can be computed by edge servers or the cloud server.
Considering the computing resource capacity, at each edge server, a tiny neural network (TNN) model (i.e., a lightweight compressed DNN model) \cite{a02} is deployed to reduce carbon emission, at the expense of higher accuracy loss. Edge servers can use green energy, e.g., solar energy \cite{a06}, and thus the mixture of energy supplied to edge servers varies from site to site caused by the volatility of green energy \cite{a08}. As a result, the carbon intensity of such mixed energy fluctuates over time and across locations \cite{b01,b01b01}.
At the cloud server, an uncompressed DNN model is deployed to achieve lower accuracy loss at the cost of higher carbon emissions.
Usually, the cloud server is more carbon-intensive than edge servers.
Clearly, total carbon emission in this framework can not exceed the purchased CER, thus the cost of purchasing CER should be below the operator's budget (e.g., pre-defined hourly or daily).
Therefore, it is {{imperative}} to design a proper policy to coordinate the two methods (i.e., on-demanded and on-preserved) to control the cost of purchasing CER. Besides, through ML tasks offloading between devices to servers, the performance (i.e., carbon emission and accuracy loss) of inference services will be enhanced.

Given the above intuition, we design carbon-aware ML task offloading with the CER purchasing budget to achieve the minimization of all ML tasks' inference accuracy loss in a long-running time.
Therefore, here comes our challenge: for the collaborative edge computing inference tasks, how to offload each of them (e.g., to which edge server or to the cloud server) and how to use diversified CER purchasing methods to minimize the total inference accuracy loss under the long-term time-averaged CER purchasing budget.

Unfortunately, it is tough to tackle such a problem.
On the one hand, the accuracy loss with the cost of purchasing CER should be carefully weighed in an efficient manner, due to the dilemma that offloading ML tasks to edge servers may emit less carbon dioxide with the cost of higher accuracy loss, while choosing the cloud server may get better performance in inference accuracy at the expense of carbon emission.
On the other hand, coordinating various ways of purchasing CER at multiple timescales is challenging, due to the fact that it is cheaper to on-preserved purchase CER but requires future information; in contrast, purchasing CER on-demanded is in a real-time manner to support the operator but with the higher price. What is more, the uncertainty and fluctuating on prices of CER bring a plethora of difficulties in obeying the long-term time-averaged CER purchasing budget.

To cope with these challenges and achieve green edge AI, we will make carbon-aware ML task offloading decisions and CER purchasing decisions to minimize accuracy loss under the CER purchasing budget only using current system statistics.
To the best of our knowledge, our work is a low-carbon and high-accuracy early effort to green edge AI by harnessing spot and future carbon markets that optimizes ML tasks in the collaborative edge computing setup with an offloading function, facing the uncertainty and fluctuating of the carbon intensity in edge servers and CER purchasing prices within the constraint of a long time average CER procurement cost budget.
The contributions are summarized below:
\begin{enumerate}
    \item \textbf{Carbon-aware ML task offloading model}: We propose a carbon-aware ML task offloading model in a collaborative edge computing environment, and formulate the long-term problem for low-carbon and high-accuracy inference services for ML tasks to achieve greening edge AI. Our goal is to minimize the total inference accuracy loss by making proper offloading decisions as well as efficient CER purchasing policies on the fly, subject to a long-term time-averaged CER purchasing budget.
    \item \textbf{An online algorithm in multi-timescale to deal with coupled formulation}: We extend the two-timescale Lyapunov optimization technique \cite{p1} to propose an efficient approximate dynamic optimization scheme of the two-timescale online algorithm (TTOA). It can properly coordinate with the multiple ways of CER purchasing through spot and future carbon markets and the carbon-aware ML tasks offloading in the collaborative edge computing framework.
    Considering the NP-hardness of the $T$-slot problems, we further propose the resource-restricted randomized dependent rounding algorithm (R3DRA) to help to gain the near-optimal solution.
    \item \textbf{Detailed theoretical analysis and trace-driven performance evaluation:}
    The detailed theoretical analysis gives the rounding gap of R3DRA which is proportional to the optimal value, and TTOA achieves a fine-tuned trade-off between the accuracy loss and the CER purchasing cost.
    Extensive simulations show the excellent performance of our algorithms. Specifically, compared with the non-trivial baselines, our TTOA can achieve up to $57.3\%$ cost of purchase cost reduction with at most $3\%$ performance loss incurred based on the real carbon intensity trace.
\end{enumerate}
The rest of this article is structured as follows.
Related work of machine learning (ML) tasks offloading and CER purchasing is presented in Sec. II and the long-term problem of low-carbon and high-accuracy inference services for ML task offloading is shown in Sec. III. In Sec. IV, we give the illustration of the approximated dynamic optimization algorithm. The performance analysis including the rounding gap of optimal value and the achieved bound is presented in Sec. V. Performance evaluation driven by the real carbon intensity trace is performed in Sec. VI while the conclusion is illustrated in Sec. VII.

\section{Related work}
There are many works toward achieving high inference accuracy in ML tasks offloading \cite{a09,a10,a11,a12}.
Yang et al. \cite{a09} studied accurate live video analytics by introducing the coordination framework of end-edge-cloud with the aim of low latency and accurate analytics.
To achieve high-quality and real-time performance through the synergy of device-edge, Li et al. \cite{a10} proposed a framework named \textbf{Edgent} which contains DNN partitioning and right-sizing.
Wu et al. \cite{a11} focus on studying the sampling rate adaption.
By exploiting deep reinforcement learning, Zhang et al. \cite{a12} proposed a new scheme of resource management to efficiently improve inference accuracy.
{{The above-mentioned works enhance the inference performance but lack consideration of the varies of the CER purchasing price and the carbon intensity of edge servers and the cloud server.}}

Orthogonally, low-carbon computing has been widely studied to mitigate the problem of global warming caused by increasing carbon emissions.
To reduce the carbon emission in a scheduling policy, Yang et al. \cite{a18} devoted to leveraging Lyapunov optimization to adapt to renewable energy's nature of variable and unpredictable and design a scheduling policy based on carbon intensity.
In \cite{a20}, Yu et al. studied the minimization of the carbon footprint of the task offloading oriented by carbon footprint considering energy sharing as well as battery charging.
{{Savazzi et al. \cite{a19} aimed at designing green federated learning which is distributed and the proposed framework can quantify the energy footprints and the carbon equivalent emissions.}}
Lannelongue et al. \cite{a21} researched a methodological framework to help welly achieve the estimation of the computational task's carbon footprint standardized and reliably.
{{Works of \cite{a18} and \cite{a20} mainly focused on exploring the advantages of renewable energy and may be overly dependent on green energy, while works of \cite{a19} and \cite{a21} looked into the quantization of carbon footprints, the results of which may be less accurate and fail to be used in reality.}}

Carbon emission rights is proposed as a tradable commodity to achieve emission reduction target \cite{a13}. and the research efforts in this area are increasing.
Guo et al. \cite{a16} researched the comparison of trading generation lights and carbon emission and found a correlation between these two.
{{To exploit the determinants of the carbon emission price, Guan et al. \cite{a17} conducted an empirical analysis based on the VEC model. }}
These studies, however, the first look of which is not buyers.
From the perspective of buyers, i.e., operators, these works focused on the minimization of CER purchasing costs while achieving better service performance.
In \cite{a14}, Wu et al. studied the microgrid's market model to trade electricity and CER while ensuring the total CER constraint.
Zhang et al. \cite{a15} devoted to studying the closed-loop supply chain network's equilibrium decision.
{{These efforts fall short of exploring the advantages of the lower price in the carbon future market and the convenience in the spot market.}}

\section{System model and problem formulation}
\subsection{System Model}
The considered situation is shown in Fig. 1, where multiple Internet of Things (IoT) devices are offloading their machine learning (ML) tasks to edge servers and the cloud server.
In analogy with the assumption in work \cite{i1}\cite{8815809}, edge servers may have multiple energy supplies including the power of grid and green. Thus carbon intensity in edge servers may be cheaper and lower than in the cloud server and changes dynamically because it relatives to the weather.
According to the resource capacity of edge servers, we assume that there is a tiny neural network (TNN) model maintained at each edge server to process ML tasks, and a deep neural network (DNN) model is placed in the cloud server to achieve higher inference accuracy.
Intuitively, processing ML tasks at the cloud server may incur higher carbon emissions due to the high carbon intensity, and high energy consumption of transmission and processing energy. Although executing ML tasks in edge servers causes less carbon emission, the accuracy loss may be higher which reduces the performance.
Besides, to reduce carbon emission, the operator has its carbon emission rights (CER) issued by the government or government agency which is not free of charge and would be bought in a timely manner (in the carbon spot market with a higher price) or in a specified period (in the carbon future market with a relatively cheaper price) \footnote{https://en.wikipedia.org/wiki/Carbon emission trading}.

The operator is assumed to run the system in a discrete-time horizon $\{0,1,...,KT\}$ in which $K\in\{0,1,2,...\}$ is the number of time frames (e.g., minutes or hours) each of which contains $T(T>0)$ time slots (e.g., seconds or minutes).
Therefore, the operator can purchase CER at each time slot on demand and at the beginning of each time frame on preserved. Generally, each ML task arrives in each time slot $\tau$.
{{Similar to the study \cite{9835126}, we construct each computation-intensive ML tasks $i\in\mathcal{N}(\tau)$ ($N(\tau)$ is the set of arriving tasks at $\tau$ with $|N(\tau)| \le N_{max}$) with a two-parameter tuple, i.e., $<H_i(\tau),F_i(\tau)>$.
Specifically, $H_i(\tau)$ represents the input data size, and $F_i(\tau)$ denotes the computation workload.}}
Through respective wireless channels, devices offload their tasks to edge servers and the cloud server, here we denote $\mathcal{M}=\{0,1,2,..., M\}$ as the total offloading locations, in which 0 denotes the cloud server and others denote edge servers.
For each server, i.e., location $j\in\mathcal{M}$, we denote $A_j$ as the processed accuracy loss. Note that $A_j$ is the accuracy loss processed by the TNN model in edge server $j~(j>0)$ or by the DNN model in the cloud server $j~(j=0)$, and each location only maintains one model. As we have illustrated above, the accuracy loss achieved by the cloud server is lower than in edge servers.
Besides, $P_{j}^{E}(\tau)$ and $P^{C}(\tau)$ are denoted to illustrate the energy consumption for ML task processing (in terms of processing per input data bit) in the edge server $j$ and the cloud server. The carbon intensity of edge server $j$ is denoted as $C_{j}^{E}(\tau)$ while the carbon intensity of the cloud server is denoted as $C^{C}(\tau)$.
We use $R_{lt}(\tau)$ and $R_{rt}(\tau)$ to denote the price of CER purchased on-preserved and on-demanded accordingly.
Furthermore, we introduce $r_{lt}(t)$, $r_{rt}(\tau)$ to illustrate the multi-timescale purchased CER amount in our collaboration system, respectively. We summarize the notations used in Table I.
\begin{table}[t]
	\caption{SUMMARY OF NOTATIONS}
	\begin{center}
		\renewcommand\arraystretch{1.50}
		\begin{tabular}{p{1.0cm}p{6.35cm}}
			\hline
                \hline
    		Notation & \multicolumn{1}{c}{Description} \\
			\hline
			$K$    &   the number of time frames \\
			$T$    &   the length of each time frame\\
                $\tau$,$t$ & the indicator of each time slot\\
                $\mathcal{N}(\tau)$ & the total number of ML tasks at time slot $\tau$\\
                $i$ & the indicator of ML tasks\\
                $j$ & the indicator of offloading locations\\
                $H_i(\tau)$           & the input data size of ML task $i$ at time slot $\tau$\\
                $F_i(\tau)$           & the computation workload of ML task $i$ at time slot $\tau$\\
                $\mathcal{M}$ & the total offloading locations \\
                $A_j$   & the accuracy loss processed by location $j$\\
                $P_{j}^{E}(\tau)$ &the resource price in edge server $j$\\
                $P^{C}(\tau)$  &the resource price in cloud server\\
                $R_{lt}(\tau)$ & the price of on-preserved CER\\
                $R_{rt}(\tau)$ & the price of on-demanded CER\\
                $C_{j}^{E}(\tau)$ & the carbon intensity of edge server $j$\\
                $C^{C}(\tau)$ & the carbon intensity of cloud server\\
                $R_{bug}$ &long-term time-averaged CER purchasing budget\\
			$V$ & Lyapunov control parameter\\
			\hline
			Variable & \multicolumn{1}{c}{Definition} \\
                \hline
			$r_{lt}(t)$        & the on-preserved CER purchasing policy\\
                $r_{rt}(\tau)$        & the on-demanded CER purchasing policy\\
                $x_{ij}(\tau)$            & the ML task offloading decision\\
			\hline
                \hline
		\end{tabular}
	\end{center}\vspace{-0.5cm}
\end{table}

\subsection{Online Control Decisions of CER Purchasing and ML Task Offloading}
For each time frame $kT$~$(k=0,1,..., K-1)$, in the beginning, i.e., $t=kT$, the operator will adopt the system information including the energy consumption metrics $P_{j}^{E}(\tau)$ of each edge server $j$ and $P^{C}(\tau)$ of the cloud server, the carbon intensity $C_{j}^{E}(\tau)$ of each edge server $j$ and $C^{C}(\tau)$ of the cloud server, and the on-preserved CER purchasing price $R_{lt}^{C}(t)$, then makes the decisions of CER purchasing on-preserved for the running of $T$ real-time slots in carbon future market.
{{To avoid complicated deployment based on accurate but cost-significant forecast methods for on-preserved CER usage planning over multiple time slots in each time frame, the on-preserved CER in a time frame will be divided evenly, which means at time slot $\tau\in[t+1,t+T-1]$, there is $r_{lt}(\tau)=r_{lt}(t)/T$ preserved CER. The purchased amount of on-preserved CER $r_{lt}(t)$ can be optimized adaptively over time frames and the CER in need at each time slot can also be well supplemented by optimizing the on-demanded CER purchase.}}
After the operator gets the information mentioned above and the on-demanded CER purchasing price $R_{rt}(\tau)$, it decides how much amount of CER $r_{rt}(\tau)$ purchased in the carbon spot market.
Note that, the price of CER at different timescales is not the same, i.e., $R_{rt}(\tau)>R_{lt}^{C}(t)$, and it is difficult to maintain the right scale of CER in advance because of the difficulty of accurately capturing future demand.

Let $x_{ij}(\tau)$ be the binary indicator of ML task offloading ($x_{ij}(\tau)=1$) the task arriving at the beginning of $\tau$ to edge server $j>0$ or the cloud server $j=0$.

In each time slot $\tau\in[t+1,t+T-1]$, for edge servers, according to the ML task $i$'s input size $H_i(\tau)$, the energy consumption is given by $P_{total}^{e}(\tau)=\sum_{i}\sum_{j}x_{ij}(\tau)H_i(\tau)P_{j}^{E}(\tau)$; while for the cloud server, the energy consumption is given by $P_{total}^{c}(\tau)=\sum_{i}x_{i0}(\tau)H_i(\tau)P^{C}(\tau)$.
As we have mentioned, we use $C_{j}^{E}(\tau)$ and $C^{C}(\tau)$ to denote the carbon intensity of edge server $j$ and the cloud server, respectively. So the carbon emission is
    \begin{equation}
    Car(\tau)=\sum_{j\in\mathcal{M}}C_{j}^{E}(t)*P_{total}^{e}(\tau)+\sum_{j\in\mathcal{M}}C^{C}(\tau)*P_{total}^{c}(\tau).
\end{equation}
And the cost of purchasing CER can be expressed as
\begin{equation}
    C(R(\tau))=r_{lt}(t)/T*R_{lt}(t)+r_{rt}(\tau)*R_{rt}(\tau).
\end{equation}
The inference accuracy loss of all ML tasks can be expressed as
\begin{equation}
    A(X(\tau))=\sum_{i}\sum_{j}x_{ij}(\tau)*A_{j}.
\end{equation}

Our objective mathematical problem \textbf{P1} is
\begin{equation}
\label{s1}
    \min \lim_{t\to\infty}\cfrac{1}{t}\sum_{\tau=0}^{t-1}  A(X(\tau)).
\end{equation}
And the constraints are shown below:
\begin{equation}
\label{s2}
    x_{ij}(\tau)\in\{0,1\}, \forall i\in\mathcal{N}(\tau), j\in\mathcal{M};
\end{equation}
\begin{equation}
\label{s3}
    \sum_{j\in\mathcal{M}}x_{ij}(\tau)=1, i\in\mathcal{N}(\tau);
\end{equation}
\begin{equation}
\label{s4}
    \sum_{i\in\mathcal{N}(\tau)}x_{ij}(\tau)*F_{i}(\tau)\leq W_{j}(\tau), \forall j\in\mathcal{M};
\end{equation}
\begin{equation}
\label{s5}
    Car(\tau)\leq r_{lt}(t)/T+r_{rt}(\tau), \forall \tau.
\end{equation}

Constraint \eqref{s2} denotes that $x_{ij}(\tau)$ is a binary variable that is one if and only if task $i$ is offloaded to location $j$ in time slot $\tau$.
Constraint \eqref{s3} indicates that each ML task can be offloaded to only one location and constraint \eqref{s4} shows that the physical resources in the edge server $j$ are restricted by $W_{j}(\tau)$. Constraint \eqref{s5} indicates that the real needed CER cannot surplus the available CER. Besides, the cost of buying CER should not be beyond the following constraint:
\begin{equation}
\label{k1}
    \lim_{t\to\infty}\cfrac{1}{t}\sum_{\tau=0}^{t-1}\mathbb{E}[C(R(\tau))]\leq R_{bug}.
\end{equation}

However, it is not easy to minimize the inference accuracy loss (\ref{s1}) under the long-term time-averaged CER purchasing budget due to the unknown future system information and the coupling influence caused by the budget.
Also, both the computing resource and CER constraints in (7) and (8) make the combinatorial nature of the ML task offloading decisions much more involved. Hence, an online algorithm is designed to coordinate the on-demanded and on-preserved CER purchasing methods in our work of minimizing problem \textbf{P1}, i.e., minimizing (\ref{s1}) subject to \eqref{s2}, \eqref{s3}, \eqref{s4}, \eqref{s5} and \eqref{k1} only using the current known information.
Moreover, an efficient approximate scheme based on resource-restricted randomized dependent rounding is integrated into the two-timescale Lyapunov optimization to gain a near-optimal solution.

\section{Approximated dynamic optimization algorithm}
\subsection{Problem Transformation}
We first apply the Lyapunov optimization technique \cite{p1} to satisfy the carbon emission rights (CER) purchasing cost constraint in (\ref{k1}). Introducing a virtual queue $Q(\tau)$ for the operator:
\begin{equation}
\label{k2}
    Q(\tau+1)=\max[Q(\tau)+C(R(\tau))-R_{bug},0].
\end{equation}
Obviously, a larger value of $Q(\tau)$ means that the cost of purchasing CER exceeded the budget $R_{bug}$.
Then we introduce the quadratic Lyapunov function
\begin{equation}
    L(\boldsymbol{\Theta}(\tau))\triangleq\cfrac{1}{2}{Q(\tau)}^2.
\end{equation}
To further preserves system stability, the $T$-slot conditional Lyapunov drift is defined as
\begin{equation}
    \Delta_{T}(\boldsymbol{\Theta}(t))\triangleq \mathbb{E}[L(\boldsymbol{\Theta}(t+T))-L(\boldsymbol{\Theta}(t))].
\end{equation}
Inspired by the Lyapunov optimization technique \cite{p1} which minimizes $T$-slot \emph{drift-plus-penalty} term, we further get:
\begin{equation}
\label{r1}
    \Delta_{T}(\boldsymbol{\Theta}(t))+V\mathbb{E}\{\sum_{\tau=t}^{t+T-1} A(X(\tau))|\boldsymbol{\Theta}(t)\},
\end{equation}
here $V\geq0$ evaluates how much attention is paid to the minimization of inference accuracy loss compared with the queue stability.
The motivation for such a control parameter is as follows: Keeping $\Delta_{T}(\boldsymbol{\Theta}(t))$ towards a lower congestion state means to bug fewer CER, and consequently leads to a larger accuracy loss. On the contrary, keeping $\{\sum_{\tau=t}^{t+T-1} A(X(\tau))|\boldsymbol{\Theta}(t)\}$ at a smaller value to help us achieve a better performance in accuracy at the cost of the bigger value of queue.
In the following, we will give Theorem 1 to further present the upper bound of \eqref{r1}.
\begin{theorem}
For any value of $\boldsymbol{\Theta}(t)$, task offloading policy ${x(\tau)}^{*}$ and CER purchasing policy $r_{lt}(\tau)^{*}$, $r_{rt}(\tau)^{*}$, the upper bound of \eqref{r1} is
\begin{equation}
\label{p1}
    \begin{split}
    &\Delta_{T}(\boldsymbol{\Theta}(t))+V\mathbb{E}\{\sum_{\tau=t}^{t+T-1} A(X(\tau))|\boldsymbol{\Theta}(t)\}\\
    &\leq B_{1}T+V\mathbb{E}\{\sum_{\tau=t}^{t+T-1} A(X(\tau))|\boldsymbol{\Theta}(t)\}\\
    &+\mathbb{E}\{\sum_{\tau=t}^{t+T-1}Q(\tau)(C(R(\tau))-R_{bug})|\boldsymbol{\Theta}(t)\},
    \end{split}
\end{equation}
where $B_{1}\triangleq({C_{max}}^{2}+{R_{bug}}^{2})/2$ and $C_{max}=\max_{t} C(R(t))$.
\end{theorem}
Due to the page limit, we leave the proof of Theorem 1 in Appendix A of \cite{Appendix}.

According to the deterministic bound shown in Theorem 1, therefore, solving our original problem equals minimizing the right hand of \eqref{r1} subject to \eqref{s2}, \eqref{s3}, \eqref{s4} and \eqref{s5}.

Unfortunately, solving this problem needs knowing future information, e.g., the resource's price and carbon intensity as well as the queue length information. Though the operator can predict future statistics, the results may be less accurate and the prediction errors may degrade the operator's achieved performance which may need to be considered. Looking deeper into the $\boldsymbol{\Theta}(t)$, it depends on the CER purchasing decisions $r_{lt}(\tau)$ and $r_{rt}(\tau)$. Hence, we propose to address the unknown information by making an approximation to set the future queue backlog values as the current value, i.e., $Q(\tau)=Q(t)$ when $\tau\in[t,t+T-1]$. After taking this approximation, the ``loosened $T$-slot drift-plus-penalty bound" is shown in the following theorem.
\begin{theorem}
Let $V>0$, and $T\geq 1$. The loosened $T$-slot drift-plus-penalty bound of Theorem 1 satisfies
\begin{equation}
\label{p2}
    \begin{split}
        &\Delta_{T}(\boldsymbol{\Theta}(t))+V\mathbb{E}\{\sum_{\tau=t}^{t+T-1} A(X(\tau))|\boldsymbol{\Theta}(t)\}\\
        &\leq B_{2}T+V\mathbb{E}\{\sum_{\tau=t}^{t+T-1} A(X(\tau))|\boldsymbol{\Theta}(t)\}\\
        &+\mathbb{E}\{\sum_{\tau=t}^{t+T-1}Q(t)(C(R(\tau))-R_{bug})|\boldsymbol{\Theta}(t)\},
    \end{split}
\end{equation}
where $B_{2}\triangleq B_{1}+{C_{max}}^{2}(T-1)/2$.
\end{theorem}
We prove Theorem 2 in Appendix B of \cite{Appendix}.

Substituting $A(X(\tau))$ and $C(R(\tau))$ into inequality (\ref{p2}), we finally get the final optimization objective \textbf{P2}
\begin{equation}
    \begin{split}
    &\min_{x(\tau),r_{lt}(t),r_{rt}(\tau)}\mathbb{E}\{\sum_{\tau=t}^{t+T-1}\sum_{i}\sum_{j}x_{ij}(\tau)*VA_{j}|\boldsymbol{\Theta}(t)\}\\
    &+\mathbb{E}\{\sum_{\tau=t}^{t+T-1}Q(t)(r_{lt}(t)/T*R_{lt}(t)+r_{rt}(\tau)*R_{rt}(\tau)\\
    &-R_{bug})|\boldsymbol{\Theta}(t)\}\\
    &\qquad \quad s.t. \quad \textit{constraints} \quad \eqref{s2}, \eqref{s3}, \eqref{s4}, \eqref{s5}.
    \end{split}
\end{equation}

In the following, we first decompose problem \textbf{P2} then design an online algorithm to deal with it.

\begin{algorithm}[t]
\begin{algorithmic}[1]
\caption{Two-Timescale Online Algorithm (TTOA)}
\STATE For each time frame $kT$~$(k\in\{0,1,..,K-1\})$, in the beginning, i.e., $t=kT$, the operator observes current system information including queue backlog $Q(t)$, resource price $[P_{j}^{E}(t),P^{C}(t)]$, carbon intensity $[C_{j}^{E}(t),C^{C}(t)]$ and CER purchasing price $R_{lt}(t)$, then makes the decisions about CER procurement $r_{lt}(t)$ and task offloading decisions $x_{ij}(t)$ through minimizing problem {\textbf{P3}}:
\begin{equation}
\begin{aligned}
&\min_{x(t),r_{lt}(t)}\sum_{i}\sum_{j}x_{ij}(t)*VA_{j}+Q(t)r_{lt}(t)/T*R_{lt}(t)\\
&\qquad \quad s.t. \quad \textit{constraints} \quad  \eqref{s2}, \eqref{s3}, \eqref{s4}, \eqref{s5}.
\end{aligned}\label{ref20}
\end{equation}\par
Obtain $x_{ij}(t)$ and $r_{lt}(t)$ by using Algorithm 2.
\STATE During each frame, i.e., $\tau \in [t+1,t+T-1]$, the operator observes current system information including queue backlog $Q(t)$, resource price $[P_{j}^{E}(\tau), P^{C}(\tau)]$, carbon intensity $[C_{j}^{E}(\tau), C^{c}(\tau]$, and CER purchasing price $R_{rt}(\tau)$, then make decisions $[x_{ij}(\tau),r_{rt}(\tau)]$ through minimizing problem {\textbf {P4}}:
\begin{equation}
\begin{aligned}
&\min_{x(\tau),r_{rt}(\tau)}\sum_{i}\sum_{j}x_{ij}(\tau)*VA_{j}+Q(t)r_{rt}(\tau)*R_{rt}(\tau)\\
&\qquad \quad s.t. \quad \textit{constraints} \quad  \eqref{s2}, \eqref{s3}, \eqref{s4}, \eqref{s5}.
\end{aligned}\label{ref21}
\end{equation}\par
Obtain $x_{ij}(\tau)$ and $r_{rt}(\tau)$ by using Algorithm 2.
\STATE Update the queue $Q(\tau)$ according to Eq. \eqref{k2}.
\end{algorithmic}
\end{algorithm}

\subsection{Two-timescale online algorithm for \textbf{P2}}
Algorithm 1 presents our algorithm explicitly.
We ignore the term $Q(t)R_{bug}$ in Eq. \eqref{ref20} and Eq. \eqref{ref21} because these terms are constants at each real-time slot and have no influence on decisions. Besides, the variable $r_{rt}(\tau)$ and the price $R_{rt}(\tau)$ are ignored in Eq. \eqref{ref20} since we assumed that the price of CER purchasing on-preserved is always lower than on-demanded. That is to say, the operator will buy enough on-preserved CER which is cheaper, and does not need to bug any on-demanded CER at time slot $t=kT$. Since each slot's $\boldsymbol{\Theta}(t)$ is already known, and the expectation is conditioned on it, the expectation of Eq. \eqref{ref20} and Eq. \eqref{ref21} can be eliminated.
As presented in Fig. 2, for each time frame $kT$~$(k\in\{0,1,..,K-1\})$, at the beginning $\tau=kT$, the operator devotes to solving problem P3 to obtain the on-preserved CER $r_{lt}(t)/T$ for each time slot in frame $kT$; while at each time slot $\tau\in[kT+1,kT+T-1]$, the operator intends to solve problem P4 to get the on-demanded CER $r_{rt}(\tau)$ at each time slot.

Unfortunately, problem \textbf{P3} and \textbf{P4} both can be proven are \emph{NP-hard} \cite{9847031}, to solve it, we will introduce an efficient approximated algorithm to get the near-optimal solution.

\subsection{Approximate Algorithm for \textbf{P3} and \textbf{P4}}
There comes the idea that once the integral constraint \eqref{s2} is relaxed into the linear constraint, the relaxed problem \textbf{P3} (\textbf{P4}) becomes the standard linear programming (LP) problem and thus using simplex method \cite{p6} or other linear programming optimization technique such as \cite{p66} to easily solve.
After getting the optimal fractional policies, we need a proper rounding method to get binary solutions while maintaining the capacity constraint \eqref{s4} and CER constraint (8). Algorithm 2 shows our rounding algorithm in detail.

We first relax the integral constraint \eqref{s2}, and obtain the relaxed-\textbf{P3} named problem $\tilde{\textbf{P3}}$ as follows:
\begin{equation}
\begin{aligned}
&\min_{x(t),r_{lt}(t)}\sum_{i}\sum_{j}x_{ij}(t)*VA_{j}+Q(t)r_{lt}(t)/T*R_{lt}(t)\\
&\qquad \qquad \quad s.t. \quad \textit{constraints} \quad \eqref{s3}, \eqref{s4}, \eqref{s5},\\
&\qquad \qquad \quad x_{ij}(t)\in[0,1], \forall i\in\mathcal{N}(\tau), j\in\mathcal{M}.
\end{aligned}\label{ref22}
\end{equation}

In the same way, problem $\tilde{\textbf{P4}}$ (relaxed-\textbf{P4}) is presented by:
\begin{equation}
\begin{aligned}
&\min_{x(\tau),r_{rt}(\tau)}\sum_{i}\sum_{j}x_{ij}(\tau)*VA_{j}+Q(t)r_{rt}(\tau)*R_{rt}(\tau)\\
&\qquad \qquad \quad s.t. \quad \textit{constraints} \quad \eqref{s3}, \eqref{s4}, \eqref{s5},\\
&\qquad \qquad \quad x_{ij}(\tau)\in[0,1], \forall i\in\mathcal{N}(\tau), j\in\mathcal{M}.
\end{aligned}\label{ref23}
\end{equation}

\begin{figure}[!t]
	\centering
	\includegraphics[width=0.96\linewidth]{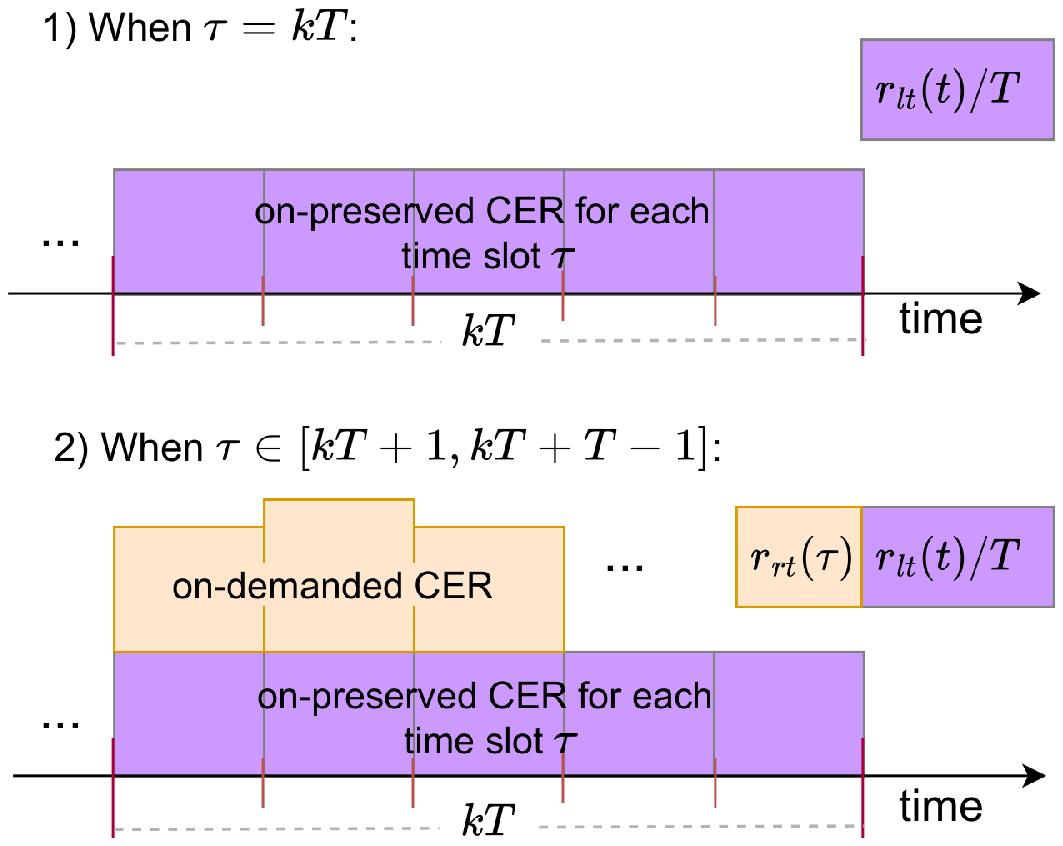}
	\caption{An example of two-timescale CER supply (where $T=5$).}\vspace{-0.45cm}
\end{figure}
Then we invoke simplex method on solving $\tilde{\textbf{P3}}$ ($\tilde{\textbf{P4}}$) thus get fractional offloading decisions $\boldsymbol{\dot{X}}$ (i.e., $\dot x_{ij}$) and purchase amount of CER decisions $\boldsymbol{\dot{R}}$ (i.e., $\dot r_{lt}$ and $\dot r_{rt}$).
For fractional offloading solutions $\boldsymbol{\dot{X}}$, a straightforward way is use Pr$[\tilde{x}_{ij}=1]=\dot{x}_{ij}$ to obtain binary decisions $\tilde{x_{ij}}$.
However, using this method will sometimes incur infeasible solutions.
For example, there may be a situation that all fractional solutions $\dot{x_{ij}}$ where $j$ equals all are rounded to 1, incurring the high cost of purchasing CER in cloud $j=0$ or violating constraint \eqref{s4} of edge servers $j>0$.
Considering this deficiency, we are encouraged to exploit the dependence of fractional solutions $\dot{x_{ij}}$ to improve optimally while obeying the computing resource and CER constraints in (7) and (8) to design a proper
resource-restricted randomized dependent rounding algorithm (R3DRA). Therefore, we will explore the dependence of decisions randomly. The key idea of R3DRA is the existence of correspondence between one rounded-up variable and another rounded-down variable, which means that there exists a compensatory between these two random variables to ensure the capacity constraint in the edge server.
\begin{algorithm}[t]
	\caption{R3DRA for Solving \textbf{P3} (\textbf{P4})}
	\hspace*{0.02in}{\bf Input:}
	The current system information;\\
	\hspace*{0.02in}{\bf Output:}
	Integral solution $\tilde{X}$ and CER purchasing decision $\tilde{R}$;
\begin{algorithmic}[1]
		\STATE Applying simplex method on solving $\tilde{\textbf{P3}}$ ($\tilde{\textbf{P4}}$) to get fractional solutions ($\boldsymbol{\dot{X}},\boldsymbol{\dot{R}}$);
		\FOR {ML task $i$}
		\STATE let $\tilde{K}_{i}^{+}=\{m|\dot{x}_{im}\in\{0,1\}\}$, $\tilde{K}_{i}^{-}=\{m|\dot{x}_{im}\in[0,1]\}$;
		\FOR {$m \in \tilde{K}_{i}^{+}$}
		\STATE set $\tilde{x}_{im}$=$\dot{x}_{im}$;
		\ENDFOR
		\FOR {$m \in \tilde{K}_{i}^{-}$}
		\STATE set  $p_{im}=\dot{x}_{im}$ and $s_{im}=F_i$;
		\ENDFOR\\
		\WHILE{$\tilde{K}_{i}^{-}>1$}
		\STATE Select $m_1$, $m_2$ from $\tilde{K}_{i}^{-}$ randomly;
		\STATE Define $\epsilon_1=\min\{1-p_{im_1},\cfrac{s_{im_2}}{s_{im_1}}p_{im_2}\}$, $\epsilon_2=\min\{p_{im_1},\cfrac{s_{im_2}}{s_{im_1}}(1-p_{im_2})\}$;
		\STATE With the probability $\cfrac{\epsilon_2}{\epsilon_1+\epsilon_2}$ set, $p_{im_1}=p_{im_1}+\epsilon_1$,  $p_{im_2}=p_{im_2}-\cfrac{s_{im_1}}{s_{im_2}}\epsilon_1$;
		\STATE With the probability $\cfrac{\epsilon_1}{\epsilon_1+\epsilon_2}$ set, $p_{im_1}=p_{im_1}-\epsilon_2$,  $p_{im_2}=p_{im_2}+\cfrac{s_{im_1}}{s_{im_2}}\epsilon_2$;
		\STATE If $p_{im_1}\in\{0,1\}$, then set $\tilde{x}_{im_1}=p_{im_1}$;
		\STATE $\tilde{K}_{i}^{+}=\tilde{K}_{i}^{+}\cup{m_1}$,  $\tilde{K}_{i}^{-}=\tilde{K}_{i}^{-}\setminus{m_1}$;
		\STATE If $p_{im_2}\in\{0,1\}$, then set $\tilde{x}_{im_2}=p_{im_2}$;
		\STATE $\tilde{K}_{i}^{+}=\tilde{K}_{i}^{+}\cup{m_2}$,  $\tilde{K}_{i}^{-}=\tilde{K}_{i}^{-}\setminus{m_2}$;
		\ENDWHILE
		\IF{$|\tilde{K}_{i}^{-}|$=1}
		\STATE Set $\tilde{x}_{im^{*}}=1$;
		\IF{constraint \eqref{s4} is broken}
		\STATE Set $\tilde{x}_{im^{*}}=0$;
		\ENDIF
		\ENDIF
		\ENDFOR
		\STATE After knowing the ML task offloading solutions $\tilde{X}$ and obtaining the purchase amount of CER decisions $\tilde{R}$, use simplex method on addressing \textbf{P3} (\textbf{P4}).
	\end{algorithmic}
\end{algorithm}

We give a simple illustration of R3DRA for solving problem $\tilde{\textbf{P3}}$ ($\tilde{\textbf{P4}}$). After we get the optimal fractional solutions $(\boldsymbol{\dot{X}},\boldsymbol{\dot{R}})$, then we denote two sets for each ML task $i$: one floating set $\tilde{K}_{i}^{-}=\{m|\dot{x}_{im}\in[0,1]\}$ and one rounded set $\tilde{K}_{i}^{+}=\{m|\dot{x}_{im}\in\{0,1\}\}$. We further denote a probability coefficient $p_{im}$ as well as a weight coefficient $s_{im}$ for each fractional $\dot{x}_{im}$. The probability coefficient $p_{im}$ is initialized as $\dot{x}_{im}$ while the weight coefficient $s_{im}$ is initialized as $F_{i}$. R3DRA runs in a series iteration to round each fractional element in set $\tilde{K}_{i}^{-}$. Specifically, in each iteration, there are two elements $m_{1}$ and $m_{2}$ randomly selected from set $\tilde{K}_{i}^{-}$, and we further introduce the coupled coefficient $\epsilon_1$ and $\epsilon_2$ in order to adjust the value of $p_{im_{1}}$ and $p_{im_{2}}$. Pay attention to that, after each adjustment, one of $p_{im_{1}}$ and $p_{im_{2}}$ is at least 0 or 1, and then we set $\tilde{x}_{im_1}=p_{im_1}$ or $\tilde{x}_{im_2}=p_{im_2}$. Therefore, in each round, fractional elements in set $\tilde{K}_{i}^{-}$ will at least decrease by 1. Eventually, when there is only one element in $\tilde{K}_{i}^{-}$, we set $p_{im}=1$ if the capacity $\sum_{i}x_{ij}F_{i}$ is not longer than $W_{j}$ and otherwise set $p_{im}=0$. As we have mentioned, the detailed expression can be seen in Algorithm 2.

{{It's worth noting that R3DRA is cost-efficient because it avoids the occurrence to offload all ML tasks to clouds to avoid incurring the high cost of purchasing CER, and is computation-efficient with the complexity only of $O(N|\mathcal{M}|)$. Therefore, the complexity of TTOA is $O(N|\mathcal{M}|KT)$.}}

\section{Performance analysis}
In the next, through rigorous analysis, we present the rounding gap caused by applying R3DRA to problem \textbf{P2} (i.e., \textbf{P3} for each frame's beginning and \textbf{P4} in each frame) and performance analysis of TTOA algorithm for problem \textbf{P1} after involving $T$-slot drift-plus-penalty methodology \cite{p1} to our objective.

Note that since we build upon the approximate solutions by R3DRA algorithm and hence the TTOA algorithm is an approximate dynamic optimization solution, our performance analysis results are different from the standard cases which require obtaining the optimal solutions for all time scales.
\subsection{Rounding Gap of R3DRA}
\begin{theorem}
By applying R3DRA to get policy $\tilde{x}_{ij}(\tau)$, $\tilde{r}_{lt}(\tau)$ and $\tilde{r}_{rt}(\tau)$, we have:
\begin{equation}
    \begin{split}
       A(\tilde{X}(\tau))+C(\tilde{R}(\tau))\leq (1+\Pi)\Phi_1H^{opt}+\Phi_2,
    \end{split}
\end{equation}
where $A(\tilde{X}(\tau))+C(\tilde{R}(\tau))$ is \textbf{P2}'s value under rounded solutions $\tilde{x}_{ij}(\tau)$ and $\tilde{r}_{lt}(\tau)$ ($\tilde{r}_{rt}(\tau)$), $H^{opt}$ is \textbf{P2}'s optimal value, $\Pi$, $\Phi_1$ and $\Phi_2$ are constants.
\end{theorem}
Due to space limits, the proof is given in Appendix C of our online technical report \cite{Appendix}. The theorem above indicates that our proposed R3RDA solution is efficient, and can guarantee an approximate solution within a bounded neighborhood of the optimal solution for problem \textbf{P2}.

\subsection{Performance Analysis of TTOA}
\begin{theorem}
After designing TTOA to deal with problem \textbf{P1} and R3DRA to address problem \textbf{P2}, the time-average accuracy loss satisfies:
\begin{equation}
\label{mhr1}
    \begin{split}
        &\sum_{t\to\infty}\cfrac{1}{t}\sum_{\tau=0}^{t-1}\mathbb{E} \{A(\boldmath{X}(\tau))\}\leq \cfrac{B_{2}}{V}+(1+\Pi)\Phi_1A^{opt}(\boldmath{X}),
    \end{split}
\end{equation}
 and the CER purchasing cost queue length satisfies:
 \begin{equation}
 \label{mhr2}
    \begin{split}
        &\sum_{t\to\infty}\cfrac{1}{t}\sum_{\tau=0}^{t-1}\mathbb{E} \{Q(\tau))\}\\
        &\leq \cfrac{B_{2}}{\eta}+\cfrac{V\Phi_1(1+\Pi)(A^{max}(X)-A^{opt}(X))}{\eta},
    \end{split}
\end{equation}
and the cost of CER purchasing is bounded:
\begin{equation}
\label{mhr3}
    \begin{split}
        &\sum_{t\to\infty}\cfrac{1}{t}\sum_{\tau=0}^{t-1}\mathbb{E}\{C(\boldmath{R}(\tau))\} \\
        &\leq \cfrac{B_{2}}{\eta}+\cfrac{V\Phi_1(1+\Pi)(A^{max}(X)-A^{opt}(X))}{\eta}+R_{bug},
    \end{split}
\end{equation}
where $A^{opt}(X)=\lim_{t\to\infty}1/{t}\sum_{\tau=0}^{t-1}\mathbb{E}\{A^{opt}(X)(\tau)\}$ is the optimal time-average accuracy loss, $A^{max}(X)$ is the largest accuracy loss, and $\eta\geq 0$ is a constant.
\end{theorem}

We prove Theorem 3 and Theorem 4 in detail in Appendix C and D of \cite{Appendix}, respectively.
From the above theorems, we can adjust the value of $V$ thus to balance the performance gap of inference accuracy well.
Next, we will give the performance evaluation driven by the real carbon intensity trace to verify our theoretical analysis.
\begin{table}[ht]
	\renewcommand{\arraystretch}{1.5}
	\caption{SIMULATION PARAMETERS}
	\begin{center}
		\begin{tabular}{lc}
			\toprule
			Parameters & Value Range\\
			\midrule
			simulation timespan             &1500\\
                {{number of offloading locations}}             &{{\{5,10,15,20\}}}\\
                {{number of tasks}}                 & {{\{5,10,20,50\}}}\\
			task input size            & $[1,10]*10^{8}$ bits\\
			the amount of computing resources  & $[5,10]*10^{11}$ cycles\\
			resource price in edge server   &$[2,5]*10^{-5}$ bit/J \\
			resource price in cloud server   & $[3,5]*10^{-4}$ bit/J\\
			computation capacity of edge server    & $[2,5]*10^{12}$ CPU cycles\\
                accuracy loss of TNN model             & $[10\%,15\%]$ \\
                accuracy loss of DNN model             & $2\%$ \\
                price of purchasing on-preserved CER             & $\mathbb{E}\{R_{lt}(\tau)\}=1.5\$$ \\
                price of purchasing on-demanded CER             & $\mathbb{E}\{R_{rt}(\tau)\}=3.0\$$ \\
                CER purchasing budget & $3.25*10^{8}$\\
			Lyapunov control parameter             &  $[1,10]*10^{8}$      \\
			\bottomrule
		\end{tabular}
	\end{center}
\end{table}

\section{Performance evaluation}
\subsection{Simulation Setup}
{{We consider there are $M\in\{5,10,15,20\}$ offloading locations consisting of $\{4,9,19,49\}$ edge servers and one cloud server and the number of ML tasks $N_{max}\in\{5,10,20,50\}$.}}
Similar to many existing studies which researched the model compressing influenced the accuracy loss such as \cite{loss}, the diverse inference accuracy loss of the TNN models of edge servers is set by randomly generating over the range $[10\%,15\%]$. In contrast, the inference accuracy loss of the DNN model in the cloud server is set as $2\%$.
{{The input data of each ML task is set as $[1,10]*10^{8}$ bits, and the workload of each ML task is set to consist of $[5,10]*10^{11}$ cycles both of them can be referenced to \cite{s1}.}}
We consider the resource price in each edge server to be in $[2,5]*10^{-5}$ bit/J while in the cloud server is in $[3,5]*10^{-4}$ bit/J both of which are similar to \cite{s2}.
And the computation capacity of edge servers is assumed to be $[2,5]*10^{12}$ CPU cycles.
The number of arriving ML tasks in the system is randomly generated over the range [1,10]. The carbon intensity of edge servers and the cloud server is according to the regional carbon intensity data in the UK (per 30 mins) of National Grid ESO \cite{carbondata}, and we use the 5 regions' data to represent the carbon intensity of edge servers and the cloud server.
We assume the price of purchasing CER $R_{lt}(\tau)$ and $R_{rt}(\tau)$ are subject to uniform random distribution and satisfy $\mathbb{E}\{R_{rt}(\tau)\}>\mathbb{E}\{R_{lt}(\tau)\}$, we also assume that $\mathbb{E}\{R_{lt}(\tau)\}=1.5\$$, $\mathbb{E}\{R_{rt}(\tau)\}=3.0\$$.
We evaluate our algorithms' performance by setting the control parameter $V\in[1,10]*10^{8}$ under the long-term time-averaged CER purchasing budget $R_{bug}=3.25*10^{8}$ cost units. The simulation runs on Matlab, where a total of $KT=1500$ (in which $T=15$ and $K=100$) randomized realizations are averaged for the current given system information.
Table II gives the simulation parameters in detail.

\begin{figure}[t]
		\subfigure[Accuracy loss vs different value of $V$.]{
    \label{Fig3(a)}
    \begin{minipage}[t]{0.49\textwidth}
	\centering
	\includegraphics[width=8.1cm]{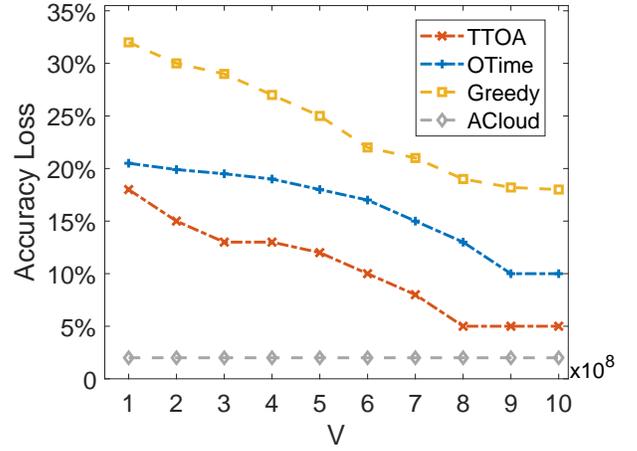}
	\end{minipage}
    }
    \centering
    \subfigure[Average queue backlog vs different value of $V$.]{
    \label{Fig3(b)}
    \begin{minipage}[t]{0.49\textwidth}
    \centering
    \includegraphics[width=8.1cm]{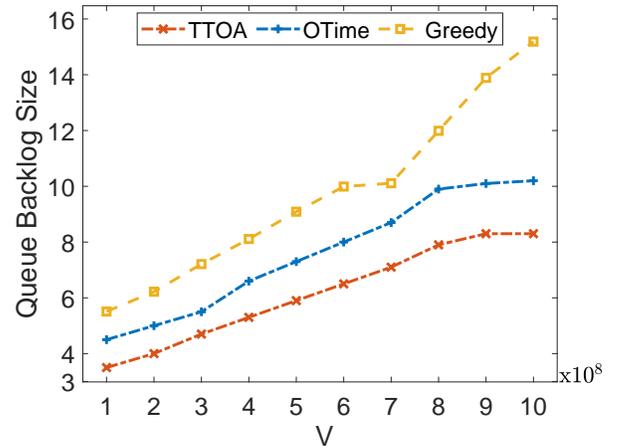} 	
    \end{minipage}
    }
    \centering
    \subfigure[Amount of offloading in TTOA vs different value of $V$.]{
    \label{Fig3(c)}
    \begin{minipage}[t]{0.49\textwidth}
	\centering
	\includegraphics[width=8.1cm]{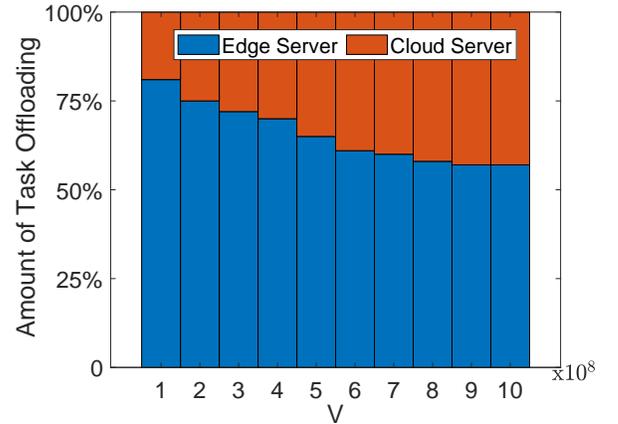}
	\end{minipage}
    }
    \centering
    \caption{The performance of TTOA.}\vspace{-0.25cm}
\end{figure}

We compare our algorithm's performance with three benchmarks which are shown below:
\begin{enumerate}
    \item All to Cloud algorithm (ACloud): Always choose the cloud server to execute ML tasks, which will achieve the best performance in inference accuracy loss but incur the highest CER purchasing cost and is an extreme contrast.
    \item One Time-scale algorithm (OTime): We zoom in on our smaller timescale, which means making our two-timescale time slot units into one. In our following experiments, the one-time slot is set as 30 minutes in this algorithm, and thus we need to decide the offloading decisions and each slot's CER purchasing amount.
    \item {Greedy algorithm (Greedy): The core idea of this algorithm is to achieve performance greedily while obeying the constraint of CER purchasing, which compares the cost of each possible offloading mode and then selects the better performance one greedily to offload the ML tasks under the CER purchasing constraint.}
\end{enumerate}

\subsection{The Accuracy Loss vs Queue Backlog trade-off}
We first show the performance of our TTOA with different values of $V$ with $T=15$ (7.5 hours a time frame). As shown in Fig. \ref{Fig3(a)}, our proposed TTOA achieves excellent performance.
For instance, it can reduce $15\%$ of the inference accuracy loss compared to Greedy and gets at most $3\%$ performance loss compared with ACloud when $V=8*10^{8}$. Besides, with the control parameter $V$ increasing, the accuracy loss will reduce, this verifies the analytical analysis in Eq. \eqref{mhr1}.
From Fig. \ref{Fig3(b)}, we can see the average queue backlog is approximately proportional to the control parameter $V$ which verifies the analytical analysis in Eq. \eqref{mhr2}, and our TTOA achieves the lowest queue backlog size. Obviously, a larger value of $V$ means that we pay more attention to minimizing accuracy loss than obeying the purchasing CER cost budget. From Fig. \ref{Fig3(c)}, it is clearly seen that with a larger value of $V$, more ML tasks will be influenced by the cloud server hence achieving lower accuracy loss in Fig. \ref{Fig3(a)} and higher queue backlog size in Fig. \ref{Fig3(b)}.

\begin{figure}[t]
    \subfigure[Cost of purchase vs different value of $V$.]{
    \label{Fig4(a)}
    \begin{minipage}[t]{0.49\textwidth}
	\centering
	\includegraphics[width=8.1cm]{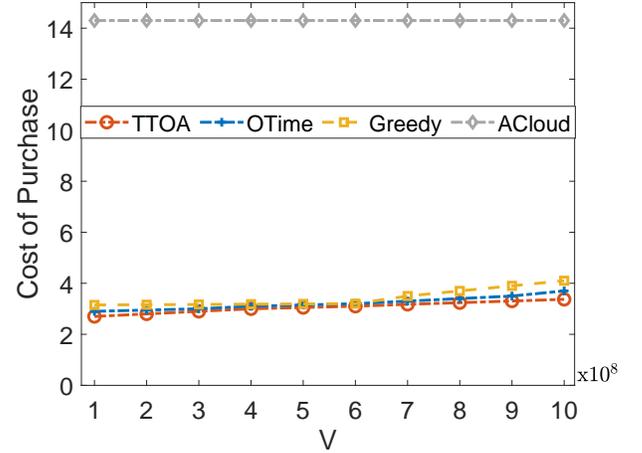}
	\end{minipage}
    }
    \centering
    \subfigure[Cost of purchase vs time.]{
    \label{Fig4(b)}
    \begin{minipage}[t]{0.49\textwidth}
    \centering
    \includegraphics[width=7.75cm]{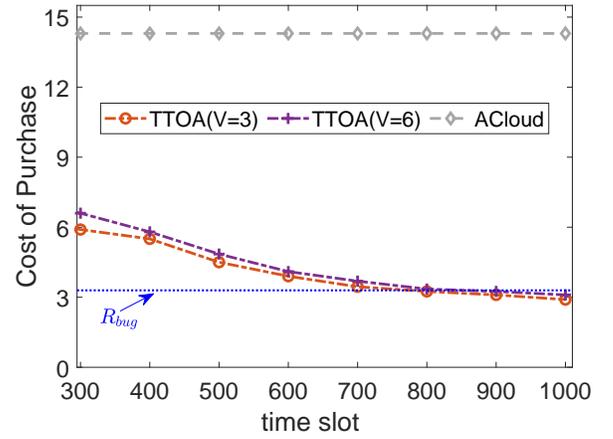}
    \end{minipage}
    }
    \centering
    \subfigure[Under Gaussian distribution of emission right price.]{
    \label{Fig4(c)}
    \begin{minipage}[t]{0.49\textwidth}
    \centering
    \includegraphics[width=8.1cm]{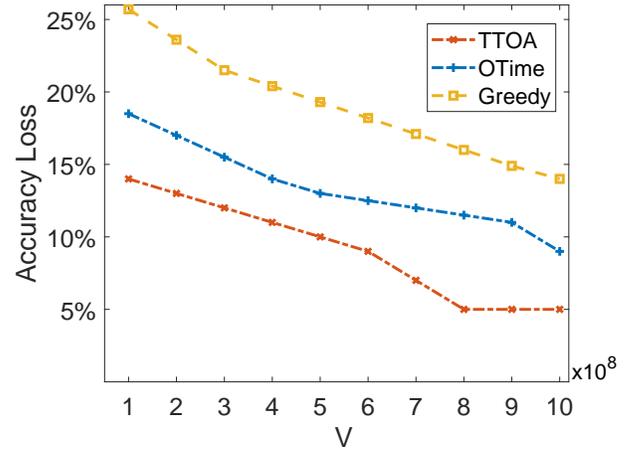}
    \end{minipage}
    }
    \centering
    \caption{The performance of TTOA.}\vspace{-0.25cm}
\end{figure}

The performance of the cost of purchase with different values of control parameter $V$ can be seen in Fig. \ref{Fig4(a)}, in which we can see the cost of purchasing CER grows with the values of $V$ increasing, and our TTOA uses the lowest cost while the ACloud causes the highest cost. For instance, when $V=8*10^{8}$, our TTOA achieves an up to $57.3\%$ cost of purchase reduction over the ACloud.
And we further evaluate the performance of the cost of purchasing CER with different time slots in Fig. \ref{Fig4(b)}. We can know that as the time slot increases, the cost of the CER purchase in our TTOA decreases and finally blows the long-term CER purchasing budget $R_{bug}$.
Therefore, the performances as shown in Fig. \ref{Fig4(a)} and Fig. \ref{Fig4(b)} have verified the analytical analysis in Eq. \eqref{mhr3}, i.e., the time-average carbon emission purchasing cost bound is roughly proportional to $V$ and lower than CER purchasing budget $R_{bug}$. It is obvious to know that our TTOA achieves the lowest carbon emission.
{{As can be seen in Fig. \ref{Fig4(c)}, to verify the robustness of our TTOA, we further evaluate the performance while the CER purchasing price is in the Gaussian distribution in which the mean emission right price equals the uniform distribution above. It is clear to know that TTOA still performs the best performance in terms of the accuracy loss of the implementation which is the same as in the uniform case above.}}

\begin{figure}[t]
\centering
    \subfigure[Average performance vs $R_{bug}$.]{
    \label{Fig5(a)}
    \begin{minipage}[t]{0.49\textwidth}
        \centering
	\includegraphics[width=8.1cm]{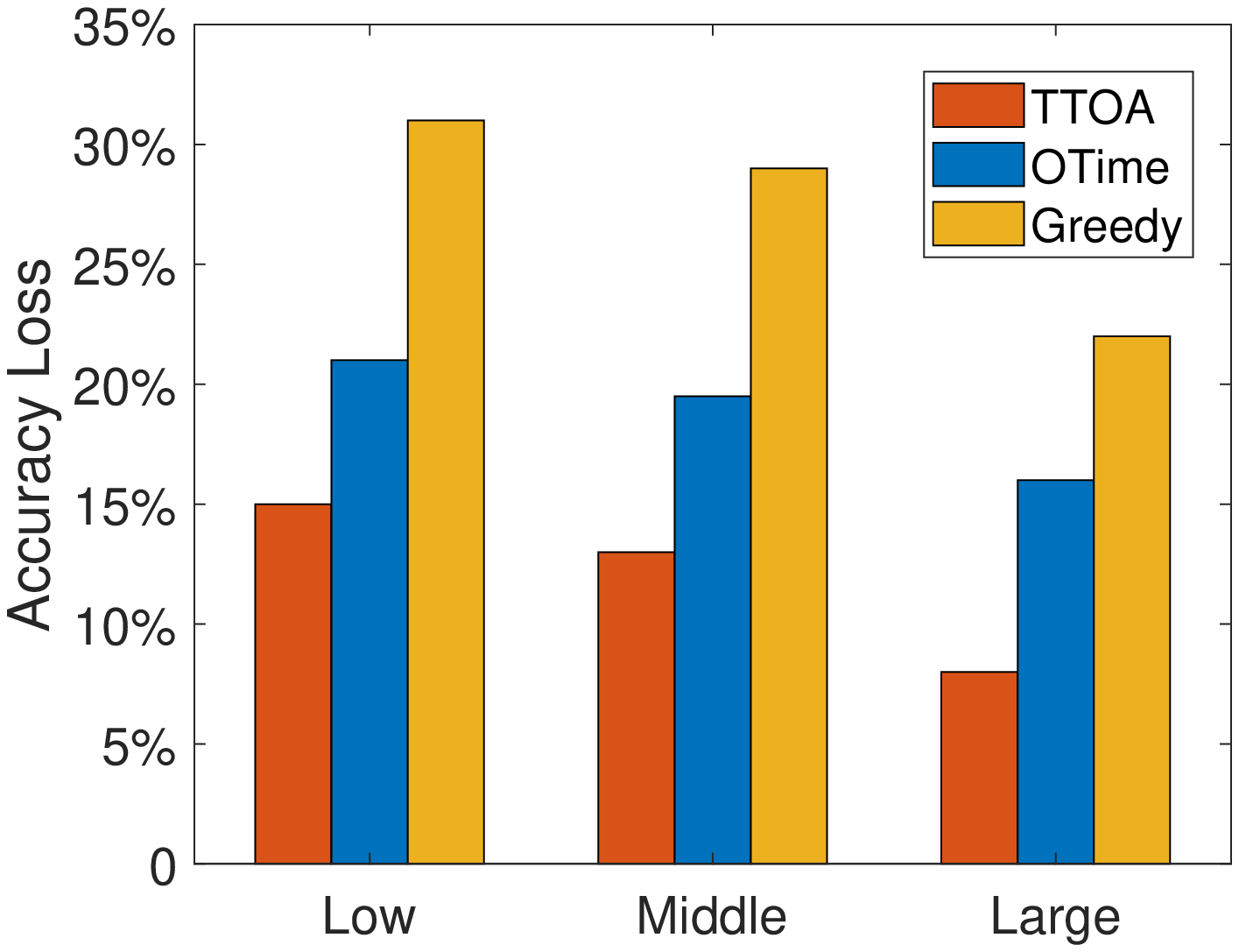}
    \end{minipage}
}
    \centering
    \subfigure[Average performance vs $T$.]{
    \label{Fig5(b)}
    \begin{minipage}[t]{0.49\textwidth}
    \centering
    \includegraphics[width=8.1cm]{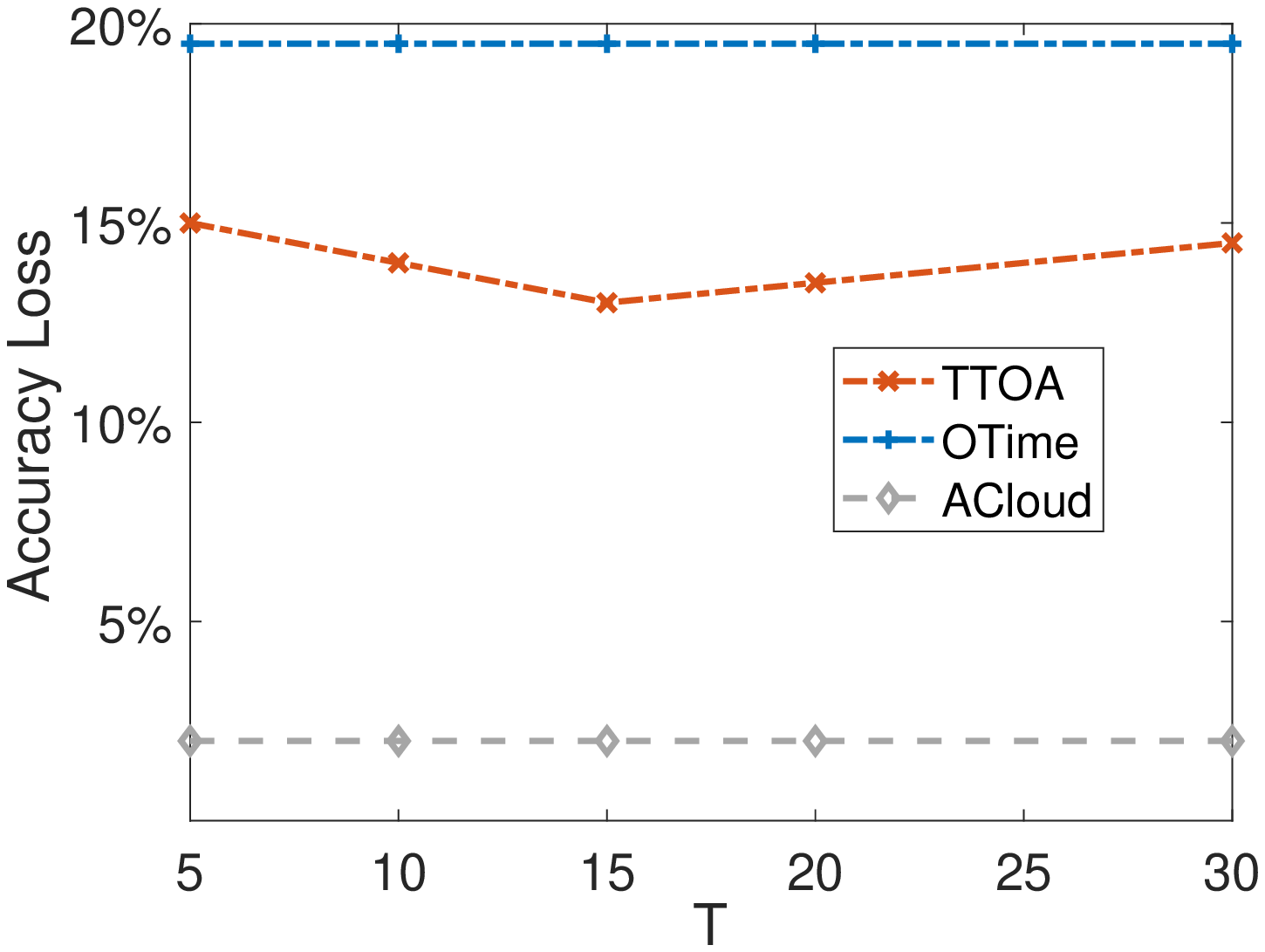}
    \end{minipage}
}
    \centering
    \subfigure[The accuracy loss and queue backlog vs $R_{lt}$.]{
    \label{Fig5(c)}
    \begin{minipage}[t]{0.49\textwidth}
        \centering
	\includegraphics[width=8.4cm]{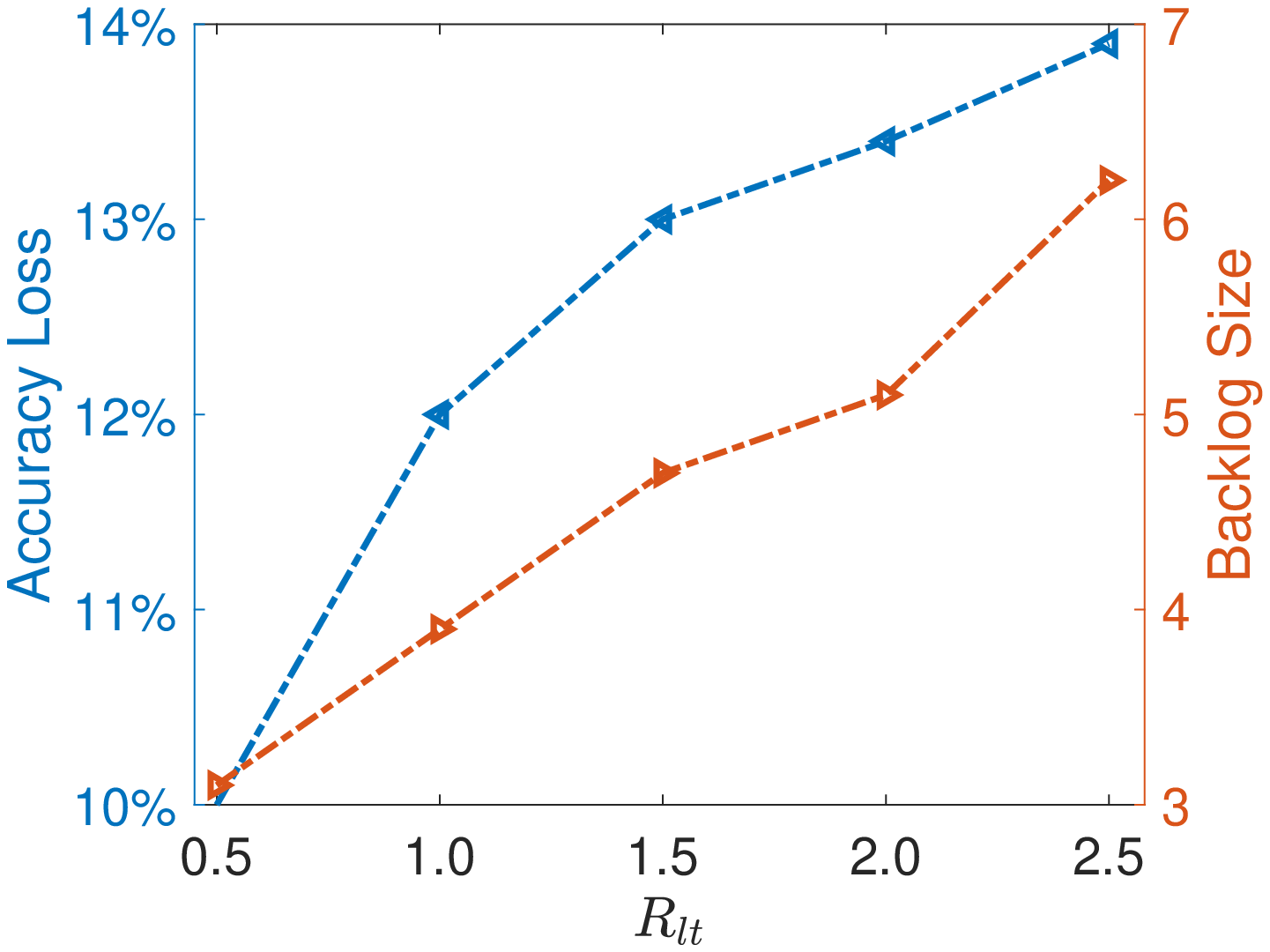}
    \end{minipage}
}
    \centering
    \caption{The impact of $R_{bug}$, $N$ and $R_{lt}$.}\vspace{-0.25cm}
\end{figure}

\subsection{Impact of budget $R_{bug}$, time frame length $T$ and CER purchasing prices $R_{lt}$ on algorithms' performance}
There is an intuition that a larger budget may supply more effort for ML tasks offloading.
In Fig. \ref{Fig5(a)}, Low, Middle, and Large are set separately as $2.25*10^8$, $3.25*10^8$, and $4.25*10^8$ cost units while the control parameter $V=3*10^8$. We see that with the long-term CER purchasing budget $R_{bug}$ increasing, compared to OTime and Greedy, our TTOA is significantly improved. For example, given the long-term CER purchasing budget $R_{bug}=3.25*10^8$ cost units, the accuracy loss of TTOA is reduced by $55.2\%$.

In Fig. \ref{Fig5(b)}, we increase the value of each time frame $T$ from 2.5 hours to 15 hours with $V=3*10^8$ (i.e., $T\in\{5,10,15,20,25,30\}$ and each single time slot is set as 30min in coordinate with the carbon intensity data trace), finding out that though the fluctuation of accuracy loss is not large, our TTOA still gains better performance than OTime. Especially, in our simulation, the accuracy loss is minimized when $T=7.5h$. The accuracy loss of the OTime is always higher than TTOA no matter what $T$'s value is. This further evaluates TTOA which contains a two-timescale and is robust to the change of $T$. Therefore though setting the value of $T$ appropriately, we can as far as possible meet the on-preserved CER period and minimize the accuracy loss.
We realize that although the values of $T$ are inappropriate, we can still make real-time CER purchasing $R_{rt(\tau)}$ to offset supply and demand nicely.

We conduct the performance of accuracy loss and queue backlog size under different CER purchasing prices $\mathbb{E}\{R_{lt}(\tau)\}$ and $\mathbb{E}\{R_{rt}(\tau)\}$ in Fig. \ref{Fig5(c)} where $\mathbb{E}\{R_{lt}(\tau)\}=0.5*\mathbb{E}\{R_{rt}(\tau)\}$. Particularly, we increase the price of $\mathbb{E}\{R_{lt}(\tau)\}$ from $0.5\$$ to $2.5\$$ (i.e., increase the price of $\mathbb{E}\{R_{rt}(\tau)\}$ from $1\$$ to $5\$$). Obviously, when the price is small, we can not only get a lower accuracy loss but also gain a lower purchase cost. Besides, when $\mathbb{E}\{R_{lt}(\tau)\}$ changes from $0.5\$$ to $1.0\$$ the accuracy loss increase seriously, while $\mathbb{E}\{R_{lt}(\tau)\}$ changes from $2.0\$$ to $2.5\$$ the queue backlog increase a lot. This is due to the limitation of the pre-defined long-term CER purchasing budget.

\begin{figure}[t]
\centering
    \subfigure[The accuracy loss vs $N_{max}$]{
    \label{Fig6(a)}
    \begin{minipage}[t]{0.49\textwidth}
        \centering
	\includegraphics[width=8.1cm]{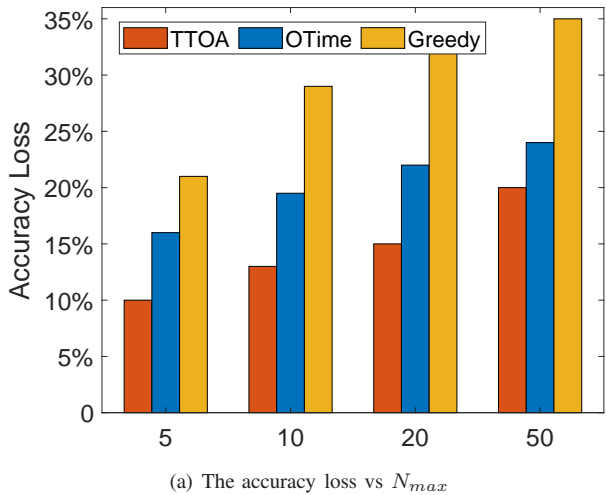}
    \end{minipage}
}
    \centering
    \subfigure[The accuracy loss vs $M$.]{
    \label{Fig6(b)}
    \begin{minipage}[t]{0.49\textwidth}
    \centering
    \includegraphics[width=8.1cm]{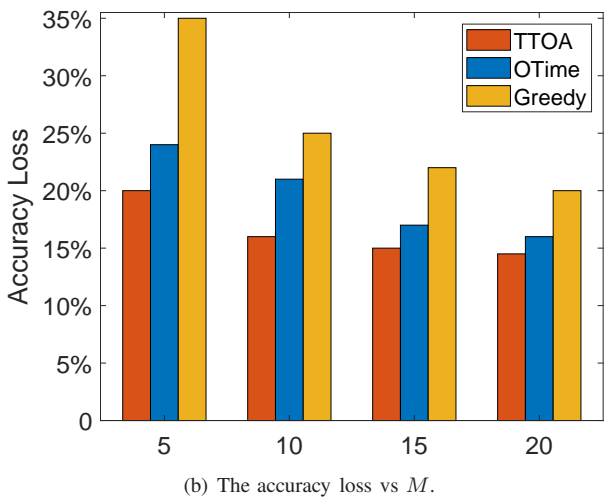}
    \end{minipage}
}
    \centering
    \caption{The scalability of TTOA.}\vspace{-0.25cm}
\end{figure}
\begin{figure}[t]
\centering
    \label{Fig7(a)}
	\includegraphics[width=8.4cm]{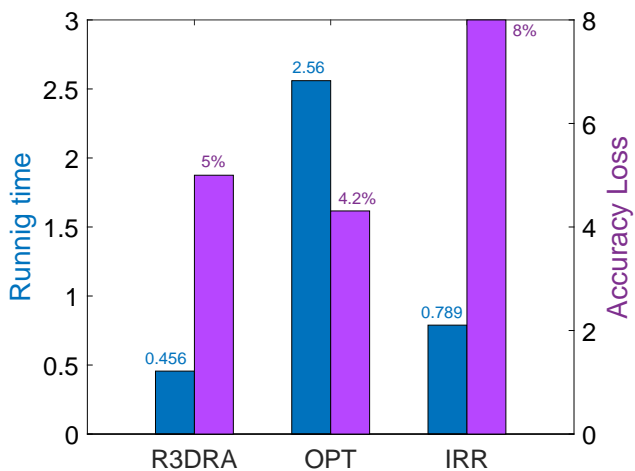}
    \caption{The performance of R3DRA.}\vspace{-0.25cm}
\end{figure}

\subsection{The impact of the number of ML tasks $N_{max}$ and the number of offloading locations $M$ on algorithms' performance}
{{As shown in Fig. \ref{Fig6(a)}, the accuracy loss in different algorithms with $M=5$ and the number of ML tasks $N_{max}\in\{5,10,20,50\}$ is presented.
Our TTOA always achieves the best accuracy loss under different values of $N_{max}$, while the performance in the Greedy is the worst because it ignores the coupling influence of each time slot's decisions when making the policy.
As we have illustrated that we zoom in on our smaller timescale in OTime algorithm to make two-timescale time slot units into one. We can see that OTime's performance is not better than TTOA's all the time, this verifies the validity of our harnessing spot and future carbon markets.}}

{{To further show the scalability of our proposed TTOA, we illustrate the accuracy loss in different algorithms with $N_{max}=50$ and the number of offloading locations $M\in\{5,10,15,20\}$ in Fig. \ref{Fig6(b)}. It is clear to know that TTOA always achieves the best performance in terms of inference accuracy, and with the number of edge servers increasing, more computing resources are provided, thus the accuracy loss of all algorithms decreases. Though the computing resources of edge servers become rich, the CER purchasing budget is the same as before, this gives the reason that the accuracy loss of $M=15$ and $M=20$ in TTOA is approximate equality.}}

{{\subsection{R3DRA's performance in TTOA}
To show the performance of our proposed rounding algorithm R3DRA, we compare R3RDR with two other rounding benchmark algorithms: the first one is OPT which uses the MILP solver to obtain the policy; the other one is IRR, i.e., uses Pr$[\tilde{x}_{ij}=1]=\dot{x}_{ij}$ to obtain binary decisions $\tilde{x_{ij}}$ while ignoring the capacity of edge servers. As presented in Fig. 7, it is obvious to see that R3DRA achieves near-optimal performance in accuracy loss compared with OPT the running time of which is nearly 6 times longer than R3DRA. Besides, though IRR's running time is not long, it violates the edge servers' capacity. To sum up, our rounding algorithm R3DRA is efficient and achieves near-optimal performance.}}

 \section{Conclusion}
In this paper, we harness spot and future carbon markets for greening edge AI towards carbon-neutral edge computing.
We propose an online carbon-aware algorithm in order to minimize ML tasks inference accuracy loss under the long-term time-averaged carbon emission rights purchasing budget in a collaborative edge computing framework.
We formulate ML tasks offloading problem \textbf{P1} with the goal of optimizing inference accuracy loss under the carbon emission rights purchasing budget. We transform and loosen the original time coupling problem to get $T$-slot loosened problem \textbf{P2}. To solve it, we develop an online algorithm TTOA that aims to solve mixed-integer linear programming problems \textbf{P3} and \textbf{P4}. In TTOA, by relaxing the feasible region to the real range, we get problems $\tilde{\textbf{P3}}$ and $\tilde{\textbf{P4}}$, to solve which we design a resource-restricted randomized dependent rounding algorithm R3DRA. We finally show the superior performance of TTOA and R3DRA through rigorous theoretical analysis and detailed experiments driven by the real carbon intensity trace.

In the future, we intend to add the prediction of green energy into the whole algorithm design so as to further contribute to achieving carbon-neutral edge computing.

\bibliographystyle{IEEEtran}
\bibliography{IEEEabrv,ref}

\end{document}